\documentclass{article}
\usepackage{PRIMEarxiv}
\usepackage{multirow}
\usepackage{longtable}
\usepackage{float}
\usepackage{amsmath}
\usepackage[utf8]{inputenc} % allow utf-8 input
\usepackage[T1]{fontenc}    % use 8-bit T1 fonts
\usepackage{hyperref}       % hyperlinks
\usepackage{url}            % simple URL typesetting
\usepackage{booktabs}       % professional-quality tables
\usepackage{amsfonts}       % blackboard math symbols
\usepackage{nicefrac}       % compact symbols for 1/2, etc.
\usepackage{microtype}      % microtypography
\usepackage{lipsum}
\usepackage{fancyhdr}       % header
\usepackage{graphicx}       % graphics
\graphicspath{{media/}}     % organize your images and other figures under media/ folder
\usepackage{longtable}
\title{Feature Group Tabular Transformer: A Novel Approach to Traffic Crash Modeling and Causality Analysis
\thanks{\textit{\underline{Citation}}: 
\textbf{Authors. Title. Pages.... DOI:000000/11111.}} 
}

\author{
  Oscar Lares, Hao Zhen, Jidong J. Yang \\
  College of Engineering \\
  University of Georgia \\
  Athens, GA 30602, USA\\
  \texttt{\{Oscar.Lares, Hao.Zhen, Jidong Yang\}@uga.edu} \\
}

\begin{document}
\maketitle

\begin{abstract}
Reliable and interpretable traffic crash modeling is essential for understanding causality and improving road safety. This study introduced a novel approach to predicting crash types by utilizing a comprehensive dataset fused from multiple sources, including weather data, crash reports, high-resolution traffic information, pavement geometry, and facility characteristics. An essential part of our proposed approach was a feature group tabular transformer (FGTT) model, which organizes disparate data into meaningful feature groups, represented as tokens. These group-based tokens serve as rich semantic components, enabling effective identification of collision patterns and interpretation of causal mechanisms. The FGTT model was compared with widely used tree ensemble models, including random forest, XGBoost, and CatBoost, demonstrating better predictive performance. Furthermore, the attention heatmaps from the FGTT model revealed key influential factor interactions, providing fresh insights into the underlying causality of distinct crash types.

\end{abstract}


\maketitle

\section{Introduction}

According to the national highway traffic safety administration (NHTSA), an estimated 40,990 people died in motor vehicle traffic crashes in 2023 \cite{nhtsa_813561}, highlighting the urgent need for enhanced road safety measures and policies. The inherent complexity of traffic dynamics, coupled with evasive factors, such as driver and human behavior, vehicular, and environmental conditions, makes crash modeling a challenging endeavor.   To develop effective safety interventions, it is essential to gain a deeper understanding of the intricate and multifaceted interactions among road users, roadway infrastructure, and weather conditions. These can include enhancements to geometric design and traffic control devices, the implementation of new traffic regulations, and the integration of advanced vehicle and infrastructure technologies. Collectively, these measures have the potential to significantly reduce the frequency and severity of traffic crashes.

Existing approaches to crash analysis have two critical gaps. First, traditional approaches rely on narrowly defined datasets, excluding critical variables such as detailed traffic patterns, environmental conditions, pavement characteristics, and driver behavior due to data accessibility challenges. This feature exclusion hampers the ability to model the multifaceted interactions among diverse factors.  Second, while machine learning (ML) and deep learning (DL) models have shown strong predictive capabilities, their interpretability remains limited, particularly in deriving actionable insights for policy formulation or engineering design.

To address these challenges, we present a novel approach that leverages an extensive, integrated dataset encompassing weather conditions, traffic data, road geometry, pavement conditions, and driver characteristics. By fusing data from multiple sources, we aim to capture the full spectrum of factors underlying traffic crashes. Building upon this rich dataset, we introduce a novel model, namely the feature group tabular transformer (FGTT), designed to enhance both predictive performance and interpretability in crash type modeling. The FGTT organizes original features into semantically meaningful units or groups, such as traffic, event, vehicle, driver, environmental, geometric, pavement, and contextual factors. These distinct groups are then represented as tokens and used within a transformer architecture \cite{vaswani2017attention}. This semantic grouping allows the model to capture interdependencies among different feature groups and facilitates a more nuanced understanding of the factors contributing to different crash types. By explicitly modeling the interactions among feature groups through the attention mechanism, the FGTT enhances causal analysis in crash modeling, leveraging attention heatmaps to uncover critical insights.

In summary, our contributions are twofold:

(1) We compile and utilize a multisource dataset that integrates critical variables often omitted in the existing literature. This comprehensive dataset enables a more holistic analysis of traffic crashes by  accounting for a wide range of influencing factors.

(2) We introduce the FGTT model, which improves predictive accuracy for crash types while enhancing  interpretability through the analysis of attention weights associated with semantic feature groups. This model not only outperforms state-of-the-art ensemble methods such as random forest, XGBoost, and CatBoost but also provides valuable insights into the relationships among feature groups, aiding in the understanding of potential confounding effects common in crash data. The attention heatmaps analysis highlights the dominant role of event-specific details (e.g., vehicle maneuvers and crash locations) and their interactions with the driver feature group, leading to different crash types. Additionally, it uncovers more complex feature group interactions specific to angle crashes. 

This study combines a comprehensive dataset with an innovative modeling approach to advance traffic safety analysis, providing both methodological advancements and practical insights to inform more effective road safety countermeasures and strategies.

\section{Crash modeling}
\label{sec:literature_review}

Crash collision and severity modeling has been a widely investigated topic utilizing various kinds of data. Typically, the approaches focus on modeling crash severity to identify factors mostly correlated with traffic crashes to provide effective countermeasures. However, not much emphasis has been placed on modeling and predicting crash types, as they are typically used as independent features within those studies. Kim et al. \cite{kim2006modeling} conducted a study to model and predict crash types at intersections in rural Georgia. Their results highlight that countermeasures targeting crash severity might only address a subset of crashes.  Besides crash severity, more effective countermeasures could be developed by understanding contributing factors to different crash types. The authors also note that crash types are associated with road geometry, environment, and traffic variables in different ways than crash severities are, which serve as justification for modeling them separately. While this study focuses on predicting the crash types, it is imperative to acknowledge the interconnection of crash types and severity. Typically, the methods employed for predicting these outcomes (crash severity vs. crash type) often overlap, leveraging similar datasets and analytical frameworks. Many crash severity studies and modeling approaches directly use the crash type as an input feature for modeling severity and risk \cite{assi2020traffic,kadilar2016effect,zeng2020investigating}. Moreover, advanced driver-assistance systems (ADAS) and other interventions can also reduce the likelihood of certain types of collisions occurring in the first place \cite{spicer2021effectiveness}. Therefore, while the primary objective of this study focuses on crash types, a comprehensive literature review pertaining to predicting crash severity is also included.

Statistical models have widely been applied to analyze crash collision and risk with respect to varying features, offering a robust framework, and identifying key risk factors. Statistical methods offer some advantages when it comes to crash modeling due to their elegant forms and ease in modeling and interpretation. However, they are limited in adequately modeling complex relationships present within crash data. For example, they often assume explicit or fixed interactions among variables, potentially oversimplifying crash complexity and disregarding the dynamic interactions of variables, such as driver behavior, the environment, and traffic. 

Additionally, the quality and availability of data can also significantly impact the model’s accuracy and reliability, with issues such as missing data and class imbalance posing notable challenges. Zeng et~al.~\cite{zeng2020investigating} investigated the effects of real-time weather conditions and roadway geometry on the severity of freeway crashes. They utilized an ordered logit model due to the ordered nature of crash severity (low, medium, high). The variables included the hourly wind speed, temperature, precipitation, visibility, and humidity alongside the horizontal curvature and grading of the roadway, and other crash- and driver- related features. Their results indicated that heavier precipitation contributed more to medium severity crashes, and that more severe crashes tended to occur on freeway segments with a small horizontal curve radius and higher vertical gradients. However, they failed to incorporate essential traffic-related features, such as speed and traffic volume, which play a crucial role in crash types and severity \cite{ahmed2011viability,hoye2020traffic}. They acknowledged the fact that higher-resolution weather data could give additional insights into the crash severity outcomes. 

Truck traffic impacts on crash severity have also been studied, since crashes involving trucks tend to be more severe due to the weight and size of the trucks involved. Xu et al. \cite{xu2023exploring} looked at the effect of truck traffic characteristics on crash severity, leveraging weigh-in-motion (WIM) sensor data over a 5-year period. They utilized both vehicle-specific and summary data, along with geometric roadway features, to reveal a correlation between mean vehicle weight and annual average daily traffic (AADT), and an increased risk of injuries and fatalities. Dutta and Fontaine \cite{dutta2019improving} investigated the use of continuous count station (CCS) sensor data, including traffic speed and volume metrics, to improve crash prediction modeling. Their findings highlighted the significance of three time scales of those features (15-minute, hourly, yearly traffic data) and their impacts on model accuracy. Other studies have explored the influence of various features on truck traffic crash severity, such as environmental conditions, vehicle and driver characteristics, roadway geometry, and crash-specific features \cite{avelar2018comparative, bedard2002independent, iranitalab2017comparison,naik2016weather}. 

Aside from classic statistical approaches, various machine learning techniques have been employed for crash severity analysis and prediction \cite{iranitalab2017comparison,santos2022literature}. While statistical models often struggle to capture the intricate and non-linear relationships in crash and traffic data due to their inherent constraints, machine learning approaches offer a viable alternative. These methods excel in modeling complex relationships and handling high-dimensional feature spaces. Assi \cite{assi2020traffic} utilized a hybrid approach comprised of principal component analysis (PCA) and support vector machines (SVM) as well as a multilayer-perceptron (MLP) using crash, roadway, driver, vehicle, and environmental characteristics to predict the severity of crashes. An increase in model performance and accuracy was noted when incorporating PCA features into the model over the initial features due to the reduction in dimensionality, however this study did not consider any traffic-related variables (e.g., speed or volume) which are known to influence the crash severity outcome. 

Morris and Yang (2020) \cite{morris2020understanding} conducted a comprehensive study on crash collision patterns on interstates in Georgia, leveraging linear discriminate analysis and extreme gradient boosting (XGBoost) to classify multi-vehicle crash collisions. Their comprehensive dataset included a wide range  of variables, encompassing roadway, traffic, weather, environmental, and driver-related features. However, the final dataset size was relatively limited (approximately 3700 crash instances) and did not consider any pavement characteristics. In a separate study, Morris and Yang \cite{morris2021effectiveness} presented an approach for addressing imbalanced data and analyzing the outcomes using three different machine learning methods. The imbalance inherently present in crash data can pose an issue for modeling due to the biases imposed as the minority classes can easily be overpowered by the majority class, resulting in poor prediction for the minority classes. To cope with the class imbalance, they investigated various resampling methods to balance the dataset more evenly using techniques like the synthetic minority oversampling technique (SMOTE) and adaptive synthetic sampling approach for imbalanced learning, and leveraging ensemble methods for prediction (CatBoost, XGBoost, random forest). Their results demonstrated that resampling methods enhanced crash collision predictions across all the models evaluated.

Deep learning methods have also been applied to crash modeling, leveraging their innate ability to handle large, high-dimensional datasets and to learn complex features and intricate relationships. However, these models can be computationally intensive and are susceptible to overfitting without careful tuning and proper regularization. While these models can deliver exceptional results, they often face the ``black box problem'', where the lack of interpretability  make it unclear why certain predictions were made. This opacity hinders the ability to derive actionable insights from the results, limiting their practical applicability \cite{geron2022hands}. Dong et al. \cite{dong2018improved} developed a deep learning model for predicting crash severity counts, utilizing a two-module architecture. The first module encodes inputs into feature representations, which are then processed by the second module for fine-tuned encoding before being passed to a regression layer to generate final crash severity predictions. The researchers evaluated their model with and without the regression layer in the final module and compared its performance to a traditional SVM model. The proposed model with the regression layer performed the best, followed by that without the regression layer, and finally the SVM. These results highlighted that even without the regression layer, the encoded feature representations significantly contributed to prediction accuracy.  The study also investigated how the model grouped closely related features (e.g., AADT, speed limit, truck percentage) into shared representation nodes, which were labeled as different representation groups based on their constituent features. Their findings indicated that geometric, pavement, and traffic feature representations had the most direct impacts on major injury crashes, minor injury crashes, and property-damage-only crashes, respectively.

Sattar et al. \cite{sattar2023transparent} also conducted a study predicting crash severity and comparing three deep learning models: an MLP, an MLP with embedding and TabNet, a popular tabular deep learning model based on transformer architectures. Their evaluation considered not only accuracy, but also precision, recall, and F1 scores, emphasizing the fact that misclassifying a severe injury is more crucial than less severe ones. Additionally, the study examined training times, given the resource-intensive nature of training deep learning models. The results showed that all three models yielded similar results for the severity classification task. However, the authors highlighted the significant differences in training times. The TabNet architecture, being more complex than the MLP models, required substantially longer training time, approximately 700\% more. TabNet has also been explored to predict the duration of traffic incidents using tabular data \cite{li2023novel}. This method demonstrated better performance compared to previous machine learning methods, excelling not only in prediction accuracy but also in interpretability. Its feature importance mechanism, driven by attention weights within the transformer architecture, enables a more intricate understanding of the model's outputs. Building on this foundation, our proposed FGTT introduces an innovative feature grouping strategy. By encoding semantically related features into tokens, the FGTT enhances both interpretability and predictive performance, offering a more transparent and effective framework for modeling traffic collisions and improving decision-making in traffic safety. 

\section{Data collection and compilation}
Multiple sources were fused together to create a comprehensive dataset. Traffic counts, pavement condition data, and crash data were sourced from the Georgia Department of Transportation (GDOT), while public weather data was readily available from online platforms. This fused dataset enabled a more nuanced understanding of crash collisions. The dataset is not only used to identify the strongly correlated features with crash types but also to provide a broader perspective on the events surrounding crash instances, thereby enhancing crash modeling in ways that other studies may not have considered. Furthermore, this study specifically focused on multi-vehicle (MV) crashes. Previous studies \cite{wang2019freeway,geedipally2010investigating} have noted differences in the factors affecting single-vehicle (SV) crashes compared to MV crashes. For instance, Wang and Feng \cite{wang2019freeway} found that factors like traffic volume and speed variance had no significant influence on SV crashes, while the opposite was true of MV crashes.

High-resolution traffic data, such as traffic counts and speeds, were obtained at 5-minute intervals from various CCSs throughout the state of Georgia. Crash incidents from police report data were first matched to these CCS stations both temporally and spatially. After the sites and crashes were matched, the high-resolution traffic data was aggregated into hourly intervals for modeling purposes. Weather data for the crash instances was obtained from  \textit{Weather Underground} \cite{WeatherUnderground}, with weather variables extracted from weather stations that were spatially matched to the corresponding CCS sites. Finally, pavement condition data was matched to the appropriate CCS sites based on geographic coordinates, providing pavement condition details for each crash instance. 

After validating and confirming the correct matches among all crashes, CCS sites, pavement features, and weather variables, the data was fused into a single combined dataset. Subsequently, feature selection and reduction were carried out. Since the crash data comes from statewide police reports, several features were purposely excluded due to their post-incident nature.  Given that this study aimed to identify the predictive variables prior to crash occurrence, after-effect features were excluded. Additionally, other crash-related features were also removed, particularly those influenced by the subjective judgment of the police officer writing the report, such as the contributing factors to the crash and the specific harmful event elements pertaining to each crash. Other redundant crash features from the police reports such as the road characteristics (e.g., whether the road was straight or curved), were omitted in favor of more accurate, quantified road and pavement measurements that better portray the road geometry and pavement conditions. 

The compiled dataset had missing values for variables such as driver ages, traffic speeds, and precipitation data. These missing values were imputed using averages from relevant groups to maintain consistency across similar conditions. For precipitation accumulation data, missing values were imputed by calculating the group mean based on the categorical variables \textit{City} and \textit{Date\_element}. The imputation was performed using a group-wise approach to preserve local weather patterns. Similarly, missing values in the \textit{Hourly\_avg\_speed} column were imputed by grouping the data based on \textit{No\_lanes}, \textit{Day\_of\_week}, \textit{Facility\_type}, \textit{Area\_type}, and \textit{Time\_of\_day}, and replacing the missing values with the mean of each respective group. This method ensured that imputations accounted for contextual variations in traffic conditions. 
After processing, the final dataset consisted of 6810 MV crash instances and 33 features. Table \ref{table:dataset_description} lists the variables analyzed in this study.

The dataset had a total of 14 numerical features and 19 categorical features. Each of the numerical and categorical feature statistics is shown in Tables \ref{table:numerical_feature_statistics} and   \ref{table:categorical_feature_distributions}. 

\begin{longtable}{lp{10cm}l}
\caption{Total dataset description.\label{table:dataset_description}} \\

\hline
\textbf{Feature} & \textbf{Description} & \textbf{Unit} \\ \hline
\endfirsthead

\hline
\textbf{Feature} & \textbf{Description} & \textbf{Unit} \\ \hline
\endhead

\hline \multicolumn{3}{r}{{\it Continued on next page}} \\ 
%\hline
\endfoot
\hline
\endlastfoot

\textit{City} & City limits where the crash took place & - \\ %\hline
\textit{Crash\_type} & Manner of collision of the car crash & - \\ %\hline
\textit{Crash\_location} & Location on the roadway where the crash took place & - \\ %\hline
\textit{Lighting} & The lighting conditions noted when the crash occurred & - \\ %\hline
\textit{Surface }& The surface roadway condition noted when the crash occurred & - \\ %\hline
\textit{Driver1\_safety\_equip} & The safety equipment in use by the primary (at-fault) driver & - \\ %\hline
\textit{Driver2\_safety\_equip} & The safety equipment in use by the secondary driver & - \\ %\hline
\textit{Veh1\_type} & Type of vehicle of the primary (at-fault) driver & - \\ %\hline
\textit{Veh2\_type} & Type of vehicle of the secondary driver & - \\ %\hline
\textit{Veh1\_maneuver} & Maneuver that the primary driver was carrying out to cause the crash & - \\ %\hline
\textit{Road\_composition} & Make-up composition of the roadway where the crash occurred & - \\ %\hline
\textit{Trafficway\_layout} & The general roadway description of how traffic flow is controlled & - \\ %\hline
\textit{Wind\_speed} & Measured wind speed at the hour closest to the crash & mph \\ %\hline
\textit{Gust} & Measured gust at the hour closest to the crash & mph \\ %\hline
\textit{Precip\_rate} & The precipitation rate for the hour closest to the crash & in/hr \\ %\hline
\textit{Precip\_accum} & The total precipitation accumulation for the day up until an hour before the crash & inches \\ %\hline
\textit{Hourly\_truck\_ratio} & Truck percentage (FHWA Classes 4--13) of total traffic in the crash direction during the preceding hour & -\\ %\hline
\textit{Hourly\_volume }& The total volume of traffic that was measured in the same roadway direction as the crash for the hour prior to the crash occurrence & count \\ %\hline
\textit{Hourly\_avg\_speed} & The average speed of traffic measured in the same roadway direction as the crash for the hour prior to the crash occurrence & mph \\ %\hline
\textit{IRI\_avg} & International Roughness Index average & in/mile \\ %\hline
\textit{Rut\_avg} & Longitudinal surface depression in the asphalt pavement of the average two-wheel paths measured between the width limits of the lane & inches \\ %\hline
\textit{Faulting\_avg\_3d} & Faulting values that fall across the entire lane & in/mile \\ %\hline
\textit{Heading\_angle} & Measures the beginning and ending of the horizontal curvature to determine the heading & degrees \\ %\hline
\textit{Percent\_grade} & Measures the grade classification of pavement sections & percentage \\ %\hline
\textit{Cross\_section\_slope} & Critical design element for pavement cross-sections & percentage \\ %\hline
\textit{Crack\_percentage} & Percentage of pavement surface exhibiting cracking & percentage \\ %\hline
\textit{Day\_of\_week} & The day of the week the crash occurred & - \\ %\hline
\textit{Driver1\_agerange} & The age range of the primary (at-fault) driver & - \\ %\hline
\textit{Driver2\_agerange} & The age range of the secondary driver & - \\ %\hline
\textit{Curvature} & The curvature classification of the roadway & - \\ %\hline
\textit{Facility\_type} & The classification type of the roadway facility & - \\ %\hline
\textit{Area\_type} & The area where the roadway facility is located, either urban or rural & - \\ %\hline
\textit{Num\_lanes }& The number of lanes of the roadway (for one direction) & count \\ %\hline
\textit{Time\_of\_day} & The time of day that the crash occurred & - \\ \hline

\end{longtable}

%\end{table}

\begin{table}[H]
\centering
\caption{Numerical feature statistics.}
\label{table:numerical_feature_statistics}
\begin{tabular}{lllll}
\hline
\textbf{Feature} & \textbf{Mean} & \textbf{Std} & \textbf{Min} & \textbf{Max} \\ \hline
\textit{Wind\_speed} & 1.22 & 1.89 & 0 & 79 \\ %\hline
\textit{Gust} & 2.32 & 3.07 & 0 & 79 \\ %\hline
\textit{Precip\_rate} & 0.01 & 0.09 & 0 & 3 \\ %\hline
\textit{Precip\_accum} & 0.10 & 0.33 & 0 & 4.52 \\ %\hline
\textit{Hourly\_truck\_ratio} & 0.08 & 0.10 & 0 & 0.88 \\ %\hline
\textit{Hourly\_volume} & 4370 & 2884 & 10 & 10,688 \\ %\hline
\textit{Hourly\_avg\_speed} & 49.06 & 15.56 & 4.18 & 80.27 \\ %\hline
\textit{IRI\_avg} & 61.43 & 33.10 & 25 & 251 \\ %\hline
\textit{Rut\_avg} & 0.107 & 0.061 & 0 & 0.41 \\ %\hline
\textit{Faulting\_avg\_3d} & 0.005 & 0.018 & 0 & 0.39 \\ %\hline
\textit{Heading} & 205 & 120.76 & 0.9 & 359.9 \\ %\hline
\textit{Percent\_grade} & -1.29 & 1.91 & -7.40 & 3.70 \\ %\hline
\textit{Cross\_section\_slope} & 0.361 & 1.691 & -4.70 & 3.4 \\ %\hline
\textit{Crack\_percentage} & 8.11 & 10.74 & 0 & 57 \\ \hline
\end{tabular}

\end{table}

\begin{longtable}{lp{12.5cm}}
\caption{Categorical feature distributions.}
\label{table:categorical_feature_distributions} \\
\hline
\textbf{Feature} & \textbf{Distribution of Categories} \\ \hline
\endfirsthead

\hline
\textbf{Feature} & \textbf{Distribution of Categories} \\ \hline
\endhead

\hline \multicolumn{2}{r}{{\it Continued on next page}} \\ %\hline
\endfoot

\hline
\endlastfoot

\textit{City} & \textit{Atlanta}: 2375 (40.16\%) \newline 
\textit{Metro Area Outside of Atlanta}: 2202 (32.33\%) \newline 
\textit{Unincorporated}: 1873 (27.5\%) \\ \hline
\textit{Crash\_location} & \textit{On Roadway - Non-Intersection}: 5513 (80.95\%) \newline 
\textit{On Roadway - Crossing/Intersection/Crosswalk/Roundabout}:~717~(10.53\%) \newline 
\textit{Entrance/Exit Ramp}: 359 (5.27\%) \newline 
\textit{Private Property/Off Roadway}: 133 (1.95\%) \newline 
\textit{Shoulder/Median/Gore}: 88 (1.29\%) \\ \hline
\textit{Lighting} & \textit{Daylight}: 5066 (74.39\%) \newline 
\textit{Dark-Lighted}: 1028 (15.09\%) \newline 
\textit{Dark-Not Lighted}: 574 (8.43\%) \newline 
\textit{Dawn}: 77 (1.13\%) \newline 
\textit{Dusk}: 65 (0.95\%) \\ \hline
\textit{Surface} & \textit{Dry}: 5465 (80.25\%) \newline 
\textit{Wet/Snow/Ice}: 1345 (19.75\%) \\ \hline
\textit{Driver1\_safety\_equip} &\textit{Lap/Shoulder Belt/Helmet Used}: 4911 (72.11\%) \newline 
\textit{Unknown}: 1,787 (26.24\%) \newline 
\textit{None Used}: 112 (1.64\%) \\ \hline
\textit{Driver2\_safety\_equip} & \textit{Lap/Shoulder Belt/Helmet Used}: 5541 (81.37\%) \newline 
\textit{Unknown}: 1208 (17.74\%) \newline 
\textit{None Used}: 61 (0.90\%) \\ %\hline
\textit{Veh1\_type} & \textit{Passenger Car/Pickup/Van/SUV}: 6060 (88.99\%) \newline 
\textit{Truck/Trailer}: 457 (6.71\%) \newline 
\textit{Unknown}: 206 (3.02\%) \newline 
\textit{Other}: 55 (0.81\%) \newline 
\textit{Motorcycle/Bicycle/ATV}: 32 (0.47\%) \\ \hline
\textit{Veh2\_type} &\textit{ Passenger Car/Pickup/Van/SUV}: 6191 (90.91\%) \newline 
\textit{Truck/Trailer}: 413 (6.06\%) \newline 
\textit{Unknown}: 142 (2.09\%) \newline 
\textit{Other}: 47 (0.69\%) \newline 
\textit{Motorcycle/Bicycle/ATV}: 17 (0.25\%) \\ \hline
\textit{Veh1\_maneuver} & \textit{Straight}: 3809 (55.93\%) \newline 
\textit{Changing Lanes/Passing}: 1886 (27.69\%) \newline 
\textit{Negotiating a Curve}: 369 (5.42\%) \newline 
\textit{Turning (left, right, u-turn)}: 285 (4.19\%) \newline 
\textit{Other}: 251 (3.69\%) \newline 
\textit{Backing}: 89 (1.31\%) \newline 
\textit{Stopped/Parked}: 85 (1.25\%) \newline 
\textit{Entering/Leaving Parking/Driveway}: 36 (0.53\%) \\ \hline
\textit{Road\_composition} & \textit{Black Top}: 5921 (86.95\%) \newline 
\textit{Concrete/Other}: 889 (13.05\%) \\ \hline
\textit{Trafficway\_layout} & \textit{Two-Way Trafficway With A Physical Barrier/Separation}: 3762 (55.28\%) \newline 
\textit{One-Way Trafficway}: 1275 (18.72\%) \newline 
\textit{Two-Way Trafficway With No Physical Barrier/Separation}: 944 (13.86\%) \newline 
\textit{Continuous Turning Lane}: 32 (0.47\%) \\ \hline
\textit{Day\_of\_week }& \textit{Monday}: 963 (14.14\%) \newline 
\textit{Tuesday}: 1096 (16.09\%) \newline 
\textit{Wednesday}: 935 (13.73\%) \newline 
\textit{Thursday}: 1135 (16.67\%) \newline 
\textit{Friday}: 1282 (18.83\%) \newline 
\textit{Saturday}: 784 (11.51\%) \newline 
\textit{Sunday}: 615 (9.03\%) \\ \hline
\textit{Driver1\_agerange} & \textit{Under 25}: 1758 (25.81\%) \newline 
\textit{25--34}: 1620 (23.79\%) \newline 
\textit{35--44}: 1856 (27.25\%) \newline 
\textit{45--54}: 723 (10.62\%) \newline
\textit{55 and up}: 853 (12.53\%)  \\ \hline
\textit{Driver2\_agerange }& \textit{Under 25}: 1244 (18.27\%) \newline 
\textit{25--34}: 1869 (27.44\%) \newline 
\textit{35--44}: 1520 (22.32\%) \newline 
\textit{45--54}: 1112 (16.33\%) \newline 
\textit{55 and up}: 1065 (15.64\%) \\ %\hline
\textit{Curvature} & \textit{A}: 5869 (86.18\%) \newline 
\textit{B}: 869 (12.76\%) \newline 
\textit{C or more}: 72 (1.06\%) \\ \hline
\textit{Facility\_type} & \textit{Interstate}: 5342 (78.44\%) \newline 
\textit{Principal Arterial - Other}: 751 (11.03\%) \newline 
\textit{Minor Arterial}: 393 (5.77\%) \newline 
\textit{Principal Arterial - Other Freeways and Expressways}: 324 (4.76\%) \\ \hline
\textit{Area\_type} &\textit{ Urban}: 6,691 (98.25\%) \newline 
\textit{Rural}: 119 (1.75\%) \\ \hline
\textit{Num\_lanes} & 
2: 1746 (25.64\%) \newline 
3: 618 (9.07\%) \newline 
4: 585 (8.59\%) \newline
5: 954 (14.01\%) \newline 
6: 709 (10.41\%)  \newline 
7: 2198 (32.28\%)\\ \hline
\textit{Time\_of\_day} & \textit{Early morning}: 428 (6.28\%) \newline
\textit{Peak morning}: 1185 (17.40\%) \newline 
\textit{Midday}: 1315 (19.31\%) \newline 
\textit{Peak afternoon}: 2857 (41.95\%) \newline 
\textit{Late evening}: 1025 (15.05\%) \\ %\hline

\end{longtable}

\section{Crash type classification}
In this study, we frame crash type inference as a multi-class classification problem.  To enhance both performance and interpretability in tabular data modeling, we propose the FGTT, leveraging the transformer architecture \cite{vaswani2017attention}. The FGTT is evaluated against three tree ensemble baselines, including random forest (RF), extreme gradient boosting (XGBoost), and categorical boosting (CatBoost). These tree ensemble methods are well-established and widely recognized for their robust performance in supervised learning tasks involving tabular data. This section provides a brief overview of each ensemble method, followed by a detailed description of our proposed FGTT.

%that have been remarkably successful being implemented in other domains \cite{gorishniy2021revisiting, khan2022transformers, padhi2021tabular, wen2023transformers}

\subsection{Baselines}

Ensemble methods, which combine multiple decision trees to make predictions, have gained significant popularity due to their flexibility and robustness in handling large tabular datasets.  Among the most widely used ensemble methods are random forest (RF) and gradient boosting.  RF is a parallel ensemble method, while gradient boosting is a sequential ensemble method. Within gradient boosting, two state-of-the-art approaches are XGBoost and CatBoost.

\subsubsection{Random forest}

Random forest is a versatile machine learning algorithm. It operates by constructing a multitude of decision trees at the training time and outputting the most common prediction (for classification) or the average prediction (for regression) from the individual trees. This helps reduce prediction variance and prevent overfitting, a common issue with single decision trees. The algorithm randomly selects subsets of both data and features at each split point, making it more robust. For classification, each tree votes for a class, and the class with the most votes becomes the final prediction. RF is particularly well-suited for predicting crash types due to its ability to handle complex, non-linear relationships between features like speed, weather conditions, and road conditions. For example, Khanum et al. \cite{khanum2023accident} and Morris and Yang \cite{morris2020understanding} both implemented RF in their approaches in modeling crash types and severity predictions, respectively. Given its effectiveness, RF was chosen as a competitive baseline for comparison in this study.

\subsubsection{XGBoost}

XGBoost is an advanced implementation of gradient boosting algorithms \cite{chen2016xgboost}. It has gained popularity in the machine learning community for its speed, performance, and versatility.  Different from the conventional first-order tree ensemble, XGBoost is a second-order method with an objective function expressed in Eq (1).
\begin{equation}
    \mathcal{L}(\phi) = \sum_{i=1}^n l(y_i, \hat{y}_i) + \sum_k \Omega(f_k),
\end{equation}
where $l(y_i, \hat{y}_i)$ is the loss function measuring the discrepancy between the predicted $\hat{y}_i$ and the actual target $y_i$, which is approximated by a second-order Taylor expansion;  $\Omega(f_k)$ denotes the regularization term, which penalizes the complexity of tree  $f_k$, typically by the number of nodes and the magnitude of leaf weights. XGBoost recursively chooses a feature split that maximizes the gain (or a reduction in loss).  In the context of crash type classification, Yang et al. \cite{yang2022predicting} utilized XGBoost for predicting crash severity and employed the Shapley additive explanation (SHAP) for model interpretation.

\subsubsection{CatBoost}

CatBoost, short for categorical boosting, is a state-of-the-art, open-source, gradient boosting library developed by Yandex \cite{prokhorenkova2018catboost}. It is specifically designed to work well with categorical data and is known for its performance, accuracy, and ease of use. CatBoost can efficiently handle categorical features through ordered target statistics while addressing the target leakage issue.  Another distinguishing feature of CatBoost is ordered boosting, a permutation-based enhancement to traditional gradient boosting. This approach addresses bias and prediction shift by utilizing independent datasets at each gradient step, a process formally described by Eq (2).
\begin{equation}
    \hat{y}_i^{(t)} = \hat{y}_i^{(t-1)} + \alpha \cdot h_t(x_i, \mathcal{D}_{\text{train}}^{<i}),
\end{equation}
where 
\begin{itemize}
    \item $\hat{y}_i^{(t)}$ is the prediction for the $i$-th instance at iteration $t$,
    \item $h_t(x_i, \mathcal{D}_{\text{train}}^{<i})$ is the model built on the training subset $\mathcal{D}_{\text{train}}^{<i}$, which includes only data points that precede $x_i$ in the current permutation,
    \item $\alpha$ is the learning rate, controlling the step size of the boosting algorithm.
\end{itemize}

In addition to the ordered boosting, CatBoost uses oblivious trees as base predictors, which are computationally efficient and lead to faster predictions while reducing the risk of overfitting. As a state-of-the-art model for tabular data with categorical features, CatBoost naturally serves as a strong baseline for crash type prediction in this study.  Previous research has successfully applied CatBoost to crash modeling, as demonstrated by Hasan et al. \cite{hasan2024application}, Li et al. \cite{li2024analyzing}, and Morris and Yang \cite{morris2021effectiveness}. For interpreting tree ensemble models, the SHAP framework, developed by Lundberg et al. \cite{lundberg2018consistent}, has been widely adopted \cite{hasan2024application, li2024analyzing, morris2021effectiveness}. SHAP values quantify the contribution of each feature to a specific prediction by comparing the model's output with and without the inclusion of that feature, offering insights into the model's decision-making process. 

\subsection{Feature group tabular transformer}

This section introduces the proposed method, the feature group tabular transformer (FGTT). We begin with a brief overview of transformers and the multi-head self-attention mechanism, followed by an explanation of our feature grouping approach.

\subsubsection{Transformer encoder}
Transformers \cite{vaswani2017attention} initially designed for sequential data, have become the architecture of choice for a wide range of natural language processing (NLP) and computer vision tasks. The core innovation of the transformer is the self-attention mechanism which enables the model to focus on relevant input tokens based on their relationships to one another. This mechanism, referred to as scaled dot-product attention, shown in Eq (3), computes the attention weight by extracting information from other tokens in the input sequence. This allows the model to effectively capture dynamic context, semantics, and relationships within the data.
\begin{equation}
    \text{Attention}(Q, K, V) = \text{Softmax}\left(\frac{Q K^T}{\sqrt{d_k}}\right) V,
\end{equation}
where $Q$ and $K$ are the query and key vectors, with the embedding dimension of $d_k$. $V$ is the value vector. The dot products of the queries and keys are computed and divided by $\sqrt{d_k}$, followed by a softmax function to obtain the weights. This process is repeated across multiple heads $n$-times and results are concatenated together. 

\begin{figure}[H]
	\centering
	\includegraphics[width=0.6\linewidth]{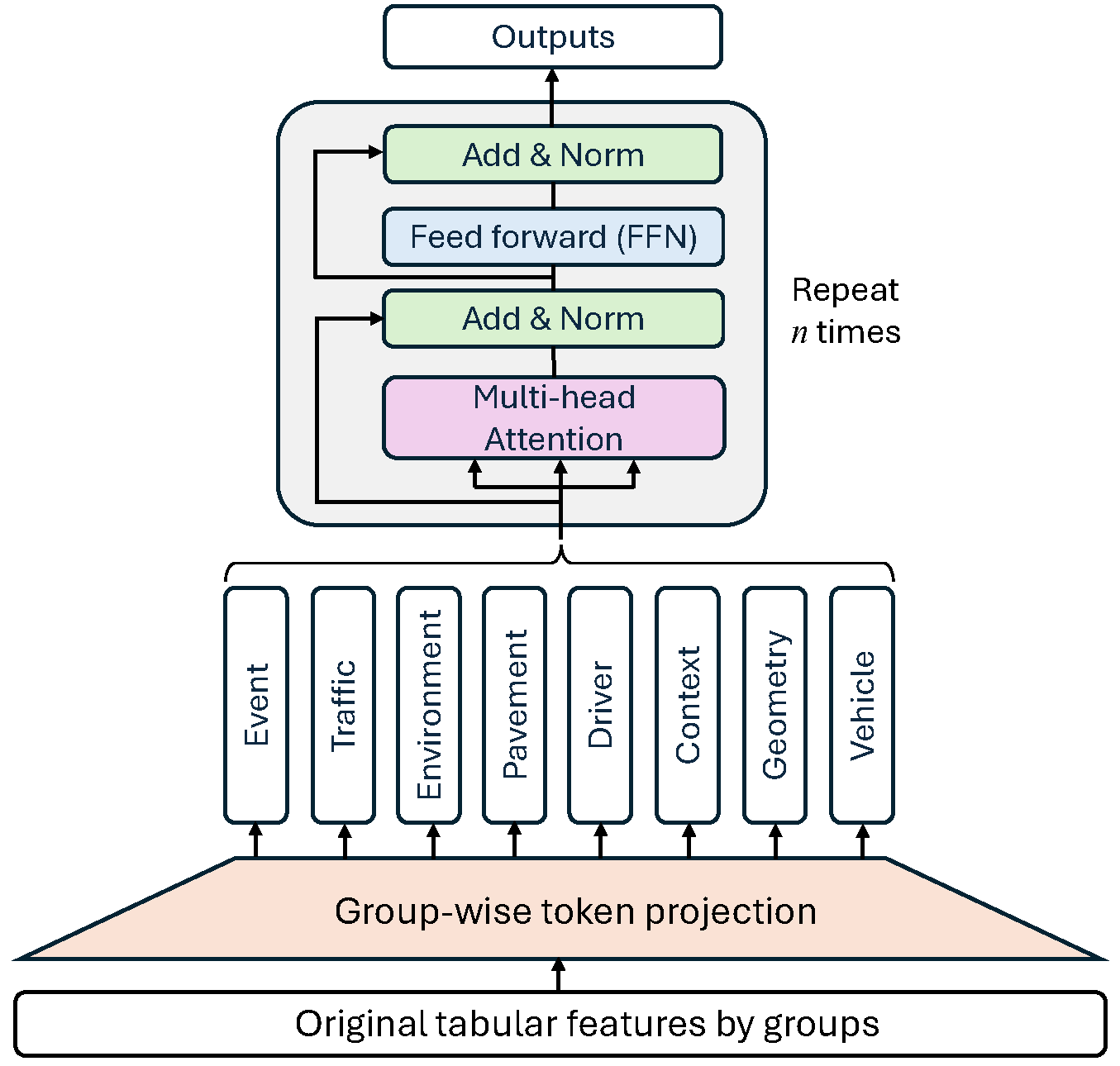}
	\caption{Transformer encoder.}
	\label{fig:transformer_encoder}
\end{figure}

In this study, only the transformer encoder is employed, as its feature encoding  is expected to enhance classification performance. The prediction approach is similar to that of Dong et al. \cite{dong2018improved}, where a linear regression head was appended to the encoder model. However, in this work, an MLP classifier was used instead of a linear regression model to make final crash type predictions \cite{sattar2023transparent}. The transformer encoder architecture utilized in this study is illustrated in Figure \ref{fig:transformer_encoder}.

\subsubsection{Feature group tokens}

Transformers are typically used in the NLP domain, where they process sequential words (or subwords) as vectorized tokens to learn context-sensitive semantic relationships between words to process sentences. Inspired by this concept, the proposed FGTT approach adapts this framework for tabular data in this study, where a \textit{crash sequence} is created, consisting of distinct tokens that encapsulate various semantic aspects of a crash event. By assigning unique semantics to each token, this approach differs from tree-based models, which treat each original feature independently.

Instead of tokenizing each individual feature in the dataset, an additional step was introduced to encode features by meaningful groups that represent distinct semantic aspects of a crash event. To achieve this, related features were grouped based on their shared characteristics and roles in determining crash outcomes. This grouping reflects different aspects of the events surrounding a crash instance while addressing the overlap between certain features.  Each feature group was then encoded into a token vector using an MLP projector, where the resulting token represented the specific semantic contribution of that feature group to the crash instance. This process produced a sequence of tokens, with each encapsulating a meaningful aspect of the crash. By organizing features into these groups, the approach aimed to provide a richer semantic representation of crashes compared to treating features independently. This richer representation was expected to better leverage the learning capability of the transformer encoder and deliver improved results over traditional ensemble methods.
  Table \ref{table:feature_groups} shows the eight feature groups defined for this study. These groups were derived from features most commonly identified in the literature as influential in determining crash outcomes.

\begin{table}[H]
\centering
\caption{Feature groups and corresponding features.}
\label{table:feature_groups}
\begin{tabular}{lp{0.3\textwidth}}
\hline
\textbf{Feature group} & \textbf{Features} \\ \hline
Event features & \textit{Crash\_location} \newline \textit{Veh1\_maneuver} \\ \hline
Traffic features & \textit{Hourly\_truck\_ratio} \newline \textit{Hourly\_volume} \newline \textit{Hourly\_avg\_speed} \\ \hline
Environment features & \textit{Gust} \newline \textit{Wind\_speed} \newline \textit{Precip\_rate} \newline \textit{Precip\_accum} \newline \textit{Lighting} \\ \hline
Pavement features & \textit{IRI\_avg} \newline \textit{Rut\_avg} \newline \textit{Faulting\_avg\_3d} \newline \textit{Percent\_grade} \newline \textit{Cross\_section\_slope} \newline \textit{Crack\_percentage} \newline \textit{Road\_composition} \newline \textit{Surface} \\ \hline
Driver features & \textit{Driver1\_AgeRange} \newline \textit{Driver2\_AgeRange} \newline \textit{Driver1\_safety\_equip} \newline \textit{Driver2\_safety\_equip} \\ \hline
Contextual features & \textit{Day\_of\_week} \newline \textit{Time\_of\_day} \newline \textit{City} \\ \hline
Geometric features & \textit{Heading\_angle} \newline \textit{Curvature} \newline \textit{Trafficway\_layout} \newline \textit{Number\_of\_lanes} \newline \textit{Facility\_type} \newline \textit{Area\_type} \\ \hline
Vehicle features & \textit{Veh1\_type} \newline \textit{Veh2\_type} \\ \hline
\end{tabular}
\end{table}

\subsubsection{Proposed FGTT model}

\begin{figure}[htbp!]
    \centering
    \includegraphics[width=0.8\linewidth]{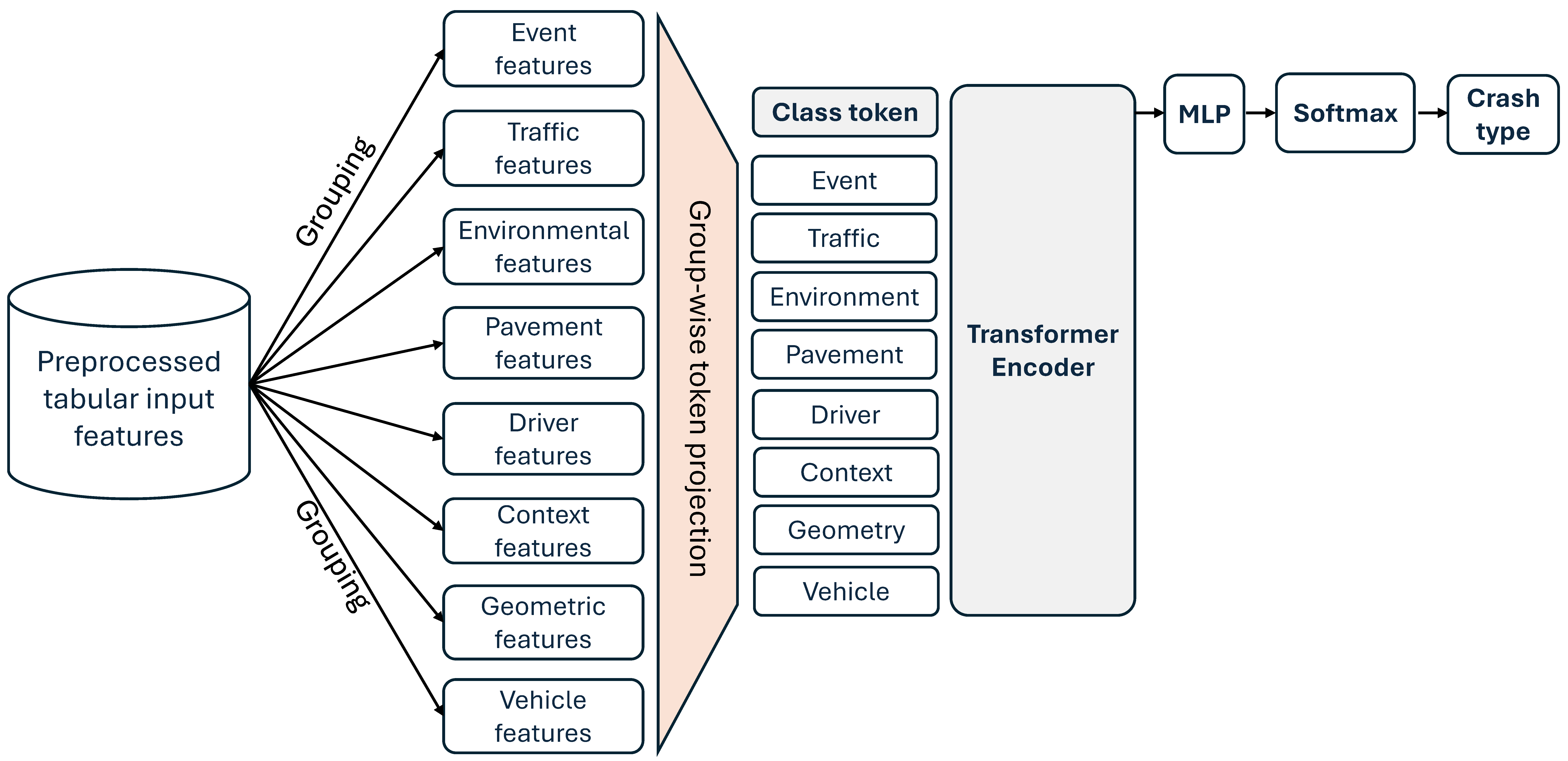}
    \caption{Illustration of the FGTT.}
    \label{fig:FGTT_model}
\end{figure}

For the proposed FGTT model, features from the dataset are first organized into distinct feature groups, each representing a specific aspect of the crash data. These groups, varying in size and dimensionality due to the differing number of features they contain, are then processed through a \textit{Feature Group Token Projection} layer. This step ensures that all feature group tokens are projected to a uniform dimension which is a necessary condition for input into the transformer encoder.  The projection process not only standardizes the input but also transforms the feature tokens into a shared representation space, enabling the model to better capture and understand the relationships and interactions among different feature groups. For this purpose, an MLP model was utilized as the feature projector. Once the tokens are projected to a common dimension, they are passed through the transformer encoder block. The encoded outputs are then fed to an MLP classifier, followed by a softmax layer to predict probabilities of crash types. The proposed FGTT model is shown in Figure \ref{fig:FGTT_model}.

\section{Experimental results and discussion}

This section outlines the experiments conducted, including the implementation of each model discussed in Section 4 and their interpretation  to better understand the possible crash causality chains. Specifically, SHAP value plots are used to explain the ensemble models, while attention weight heatmaps are generated for the FGTT model to provide deeper insights into the impact and interactions of the feature groups on crash type predictions.

Data partition and normalization are crucial steps in machine learning workflows. The original dataset, as described previously, was divided into three subsets: training, validation, and testing. Instead of using random data splitting, stratified splitting was adopted to ensure that each subset (i.e., training, validation, and testing) retains the same class distribution as the original dataset (13\% of angle crashes; 29\% of sideswipe crashes; and 58\% of rear-end crashes). This is particularly important in scenarios like this study, where certain class labels, such as angle and sideswipe crashes, are underrepresented.  As a result, the original dataset of 6810 multi-vehicle crashes was partitioned using stratified sampling based on the class label \textit{Crash type}, leading to the following splits. 

•	Training Set: 6026 samples (88.5\%);

•	Validation Set: 392 samples (5.75\%);

•	Testing Set: 392 samples (5.75\%).

Many of the features in this study are categorical. Given the small number of categories for each categorical variable, one-hot encoding is employed. To expedite and stabilize training, numeric features are standardized by subtracting the mean and divided by the standard deviation. To prevent information leakage, the mean and standard deviation are computed training dataset and then applied consistently across all three subsets.

This standardization process ensures that each numeric feature contributes equally to the distance computations, which is important for metric-oriented learning algorithms, such as transformers. Standardizing the numeric data in this study is particularly important in this study, as some numeric features, such as the \textit{Hourly\_volume}, span a large range of values. 

\subsection{Ensemble methods}

A GridSearch cross-validation (CV) strategy was employed for hyperparameter tuning of the ensemble models, including random forest, XGBoost, and CatBoost. CV is a commonly used technique in machine learning for hyperparameter tuning. It involves partitioning the training data into subsets, training the model on some subsets, and validating the model on the remaining subsets. This process is repeated multiple times, cycling through the subsets, to ensure that each subset serves as both training and validation data. The results are averaged to provide a final performance evaluation. To ensure proper class label distribution, the data subsets are stratified. GridSearch systematically explores the combinations of hyperparameters, cross-validating each combination to determine the best-performing set. The final models were trained with the optimal hyperparameters identified.

All tree ensemble models were trained using a consistent methodology, which included hyperparameter tuning through 5-fold GridSearch CV and leveraging a validation set to implement early stopping to mitigate the risk of overfitting. The hyperparameters for the 5-fold CV process for XGBoost, random forest, and CatBoost are listed in Table \ref{table:hyperparameters}.

\begin{table}[H]
\centering
\caption{Cross-validation hyperparameter tuning for XGBoost, random forest, and CatBoost (\textbf{bold} indicates the selected value).}
\label{table:hyperparameters}
\begin{tabular}{lp{13cm}}
\hline
\textbf{Model}         & \textbf{Cross-validation Parameters}                                      \\ \hline
Random Forest         & \textit{$\eta$}:\{0.01, \textbf{0.05}, 0.1, 0.3\}; \textit{n\_estimators}:\{100, \textbf{200}, 500\};
\textit{max\_depth}:\{None, 10, 20, \textbf{30}\}; \textit{min\_samples\_split}:\{2, \textbf{5}, 10\} \\ %\hline
XGBoost                & \textit{$\eta$}:\{0.01, \textbf{0.05}, 0.1, 0.3\}; \textit{n\_estimators}:\{\textbf{100}, 200, 500\}; \textit{max\_depth}:\{3, 5, \textbf{7}, 9\}\\ %\hline
CatBoost               & \textit{$\eta$}:\{0.01, \textbf{0.05}, 0.1, 0.3\}; \textit{iterations}:\{100, 200, 500, 1000, \textbf{2000}, 4000\}; \textit{depth}:\{3, 5, \textbf{7}, 9\} \\ \hline
\end{tabular}
\end{table}

\subsection{FGTT}

A different strategy was employed for tuning the parameters of the FGTT model. Bayesian optimization (BO) was used to efficiently find the optimal parameter settings. Unlike GridSearch, which exhaustively evaluates every combination of parameters, BO leverages a Gaussian process to construct a posterior distribution of the objective function based on prior  evaluations. It then uses an acquisition function to decide where to sample next. This is particularly advantageous for tuning deep learning models, where the parameter space can be extremely large, and training can be computationally intensive and time-consuming. By intelligently choosing the next set of parameters to evaluate, BO significantly reduces the time and resources required as compared to the GridSearch method. This method has proven to be efficient and effective in tuning the hyperparameters of complex models \cite{turner2021bayesian, victoria2021automatic}.  The Optuna package \cite{akiba2019optuna} was used for BO implementation.

The FGTT model was trained using focal loss, which was initially implemented for object detection tasks to address the challenge associated with class imbalance \cite{ross2017focal}. It achieves this by down-weighting the loss contribution from well-classified examples, typically belonging to majority classes, and focusing more on misclassified examples, often from minority classes. This adjustment enables the model to prioritize learning from underrepresented but significant classes, improving robustness and performance on imbalanced datasets where traditional loss functions may struggle. Focal loss has been successfully applied in various domains \cite{tian2023synergetic,liu2023attention,yu2020convolutional}, demonstrating improved prediction outcomes. The parameter settings for the FGTT are listed in Table \ref{table:fgtt_parameters}.

\begin{table}[H]
\centering
\caption{Parameter settings for the FGTT.}
\label{table:fgtt_parameters}
\begin{tabular}{lll}
\hline
\textbf{Parameter}             & \textbf{Range of Values Evaluated}  & \textbf{Optimal Value}                         \\ \hline
 \textit{learning rate}                 & [0.001, 0.1]   & 0.017       \\ 
\textit{optimizer}                      & {\textit{Adam, SGD, RMSProp}} & \textit {SGD}                                            \\ 
 \textit{FFN dimension}                 & {16, 24, 32, 64}  & 64     \\ 
 \textit{hidden dimension}             & {16, 24, 32, 64}  & 64       \\
 \textit{MLP dropout rate}            & {0.1, 0.2, 0.3, 0.4}  & 0.2   \\ 
 \textit{number of heads}                     & 2, 3, 4, 6  & 4       \\ 
\textit{number of layers}                     & 2, 3, 4, 5, 6 & 3     \\ \hline
\end{tabular}
\end{table}

\subsection{Evaluation metrics}

Evaluating machine learning models on imbalanced datasets requires metrics that capture performance beyond simple accuracy. Although overall accuracy, defined as the ratio of correct predictions to total predictions, offers a general assessment, it falls short in scenarios with uneven class distributions. In such cases, metrics like precision, recall, and their harmonic mean, the F1 score, offer a more nuanced and comprehensive evaluation.

Precision measures the proportion of correctly predicted positive instances among all predicted positives:
\begin{equation}
    \text{Precision} = \frac{TP}{TP + FP}. \tag{9}
\end{equation}

Recall quantifies the ability to identify actual positive instances among all true positives:
\begin{equation}
    \text{Recall} = \frac{TP}{TP + FN}. \tag{10}
\end{equation}

Since Precision and Recall often exhibit a trade-off, balancing them is critical. The F1 score achieves this balance by calculating their harmonic mean:
\begin{equation}
    F1 = 2 \times \frac{\text{Precision} \times \text{Recall}}{\text{Precision} + \text{Recall}}. \tag{11}
\end{equation}

Unlike the arithmetic mean, the F1 score penalizes extreme imbalances between Precision and Recall, making it suitable for tasks involving critical minority classes. In this study, where rare crash types, such as sideswipe and angle crashes, are of primary interest, the F1 score serves as a more reliable and informative metric than Accuracy.

\section{Results}

The model performance metrics are summarized in Table \ref{table:modeling_performance}. 
\begin{table}[H]
\centering
\caption{Comparison of model performance.}
\label{table:modeling_performance}
\resizebox{\textwidth}{!}{%
\begin{tabular}{lllllllll}
\hline
\textbf{Modeling Method} & \textbf{Label} & \textbf{Acc.} & \textbf{Prec.} & \textbf{Rec.} & \textbf{F1} & \textbf{Rear-end} & \textbf{Sideswipe} & \textbf{Angle} \\ \hline
\multirow{4}{*}{Random forest} & Rear-end & 93.4\% & 0.845 & 0.934 & 0.887 & 93.0\% & 6.6\% & 0.4\% \\ %\cline{2-9} 
 & Sideswipe & 77.2\% & 0.740 & 0.772 & 0.756 & 17.5\% & 77.2\% & 5.3\% \\ %\cline{2-9} 
 & Angle & 29.4\% & 0.682 & 0.294 & 0.411 & 37.3\% & 33.3\% & 29.4\% \\  %\cline{2-9} 
 & \textit{Weighted Avg} & \textit{80.4\%} & \textit{0.793} & \textit{0.804} & \textit{0.787} & -- & -- & -- \\ \hline
\multirow{4}{*}{XGBoost} & Rear-end & 92.5\% & 0.850 & 0.925 & 0.886 & 92.5\% & 5.3\% & 2.2\% \\ %\cline{2-9} 
 & Sideswipe & 79.8\% & 0.752 & 0.798 & 0.775 & 15.8\% & 79.8\% & 4.4\% \\ %\cline{2-9} 
 & Angle & 27.5\% & 0.583 & 0.275 & 0.373 & 37.3\% & 35.3\% & 27.5\% \\ %\cline{2-9} 
 & \textit{Weighted Avg} & \textit{80.4\%} & \textit{0.787} & \textit{0.804} & \textit{0.787} & -- & -- & -- \\ \hline
\multirow{4}{*}{CatBoost} & Rear-end & 91.6\% & 0.863 & 0.916 & 0.889 & 91.6\% & 5.7\% & 2.6\% \\ %\cline{2-9} 
 & Sideswipe & 74.6\% & 0.752 & 0.746 & 0.749 & 15.8\% & 74.6\% & 9.6\% \\ %\cline{2-9} 
 & Angle & 41.2\% & 0.553 & 0.412 & 0.472 & 29.4\% & 29.4\% & 41.2\% \\ %\cline{2-9} 
 & \textit{Weighted Avg} & \textit{80.1\%} & \textit{0.790} & \textit{0.801} & \textit{0.794} & -- & -- & -- \\ \hline
\multirow{4}{*}{FGTT} & Rear-end & 92.1\% & 0.867 & 0.921 & 0.893 & 92.1\% & 5.7\% & 2.2\% \\ %\cline{2-9} 
 & Sideswipe & 77.2\% & 0.733 & 0.772 & 0.752 & 17.5\% & 77.2\% & 5.3\% \\ %\cline{2-9} 
 & Angle & 39.2\% & 0.645 & 0.392 & 0.488 & 23.5\% & 37.3\% & 39.2\% \\ %\cline{2-9} 
 & \textit{Weighted Avg} & \textit{\textbf{80.9\%}} &\textit{\textbf{ 0.799}} & \textit{\textbf{0.809}} & \textit{\textbf{0.799}} & -- & -- & -- \\ \hline
\end{tabular}%
}
\end{table}

The FGTT model outperformed all ensemble models, achieving the highest overall weighted F1 score of 0.799, the highest precision of 0.799, the highest recall of 0.809, and the highest accuracy of 80.9\%. This demonstrates the FGTT's ability to deliver a more balanced performance across all crash types while maintaining strong overall predictive capability. 

The CatBoost model also showed strong performance, with a weighted F1 score of 0.794, only slightly lower than the FGTT. Notably, it excelled in predicting the minority class, Angle, achieving the highest accuracy (41.2\%) and recall (0.412) for this category. This highlights CatBoost's effectiveness in handling imbalanced datasets and identifying the minority crash type. Additionally, CatBoost exhibited high precision (0.863) and a high F1 score (0.889) for the rear-end category, showcasing its reliability for majority class predictions.

The XGBoost model, while slightly trailing in overall performance, achieved a competitive weighted F1 score of 0.787. It displayed balanced predictive performance across categories but struggled in the Angle class, with the lowest accuracy (27.5\%) and F1 score (0.373). 

The random forest model excelled in predicting the majority class (Rear-end), achieving the highest accuracy (93.4\%) and F1 score (0.887) among all models for this category. However, its performance declined for the minority class, Angle, with an accuracy of 29.4\% and an F1 score of 0.411. Despite this, the RF model maintained a competitive overall weighted F1 score of 0.787 matching XGBoost.

In summary, the FGTT delivered the best overall performance, while CatBoost demonstrated better capabilities for minority class prediction. Both models showcased the advantages of ensemble learning techniques in achieving balanced and accurate predictions across crash types. 

In addition to comparing overall model performance, further analysis and interpretation of model predictions were performed using SHAP values for the ensemble methods and attention weight heatmaps for the proposed FGTT model. 

SHAP values provided insights into feature importance by quantifying each feature's contribution to the prediction, enabling a deeper understanding of how individual features influenced the model's decisions for different crash types.  Specifically, the SHAP value for a given feature measures its contribution to the final model prediction, weighted relative to the contributions of all other features.  In this study, the SHAP values for the CatBoost model were averaged across all classes (rear-end, sideswipe, angle) since CatBoost has the best performance among the ensemble models. This analysis highlights the magnitude of each feature's impact on the model's prediction.

\subsection{Feature importance by SHAP values for CatBoost}

\begin{figure}[H]
	\centering
	\includegraphics[scale=0.35]{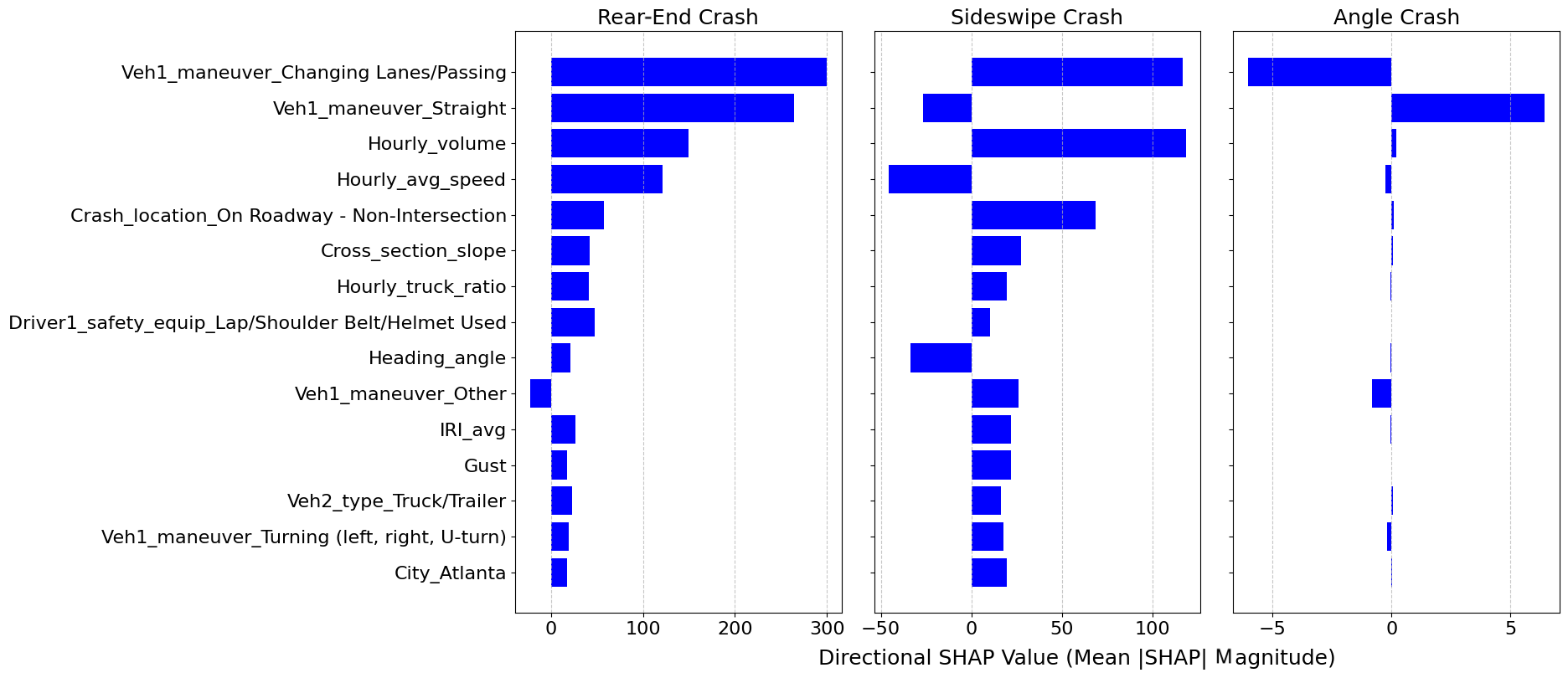}
	\caption{Directional SHAP values for CatBoost.}
	\label{fig:SHAP_cat_dir}
\end{figure}

Figure \ref{fig:SHAP_cat_dir} shows the top fifteen features based on SHAP values for the CatBoost model. The analysis highlights that the event features, such as the maneuver of vehicle 1 (\textit{Veh1\_maneuver}) and crash location, and traffic features are consistently the most significant features in inferencing crash types, as evidenced by their high SHAP values. For the event features, the maneuver of vehicle 1 such as changing lanes or passing and keeping straight all have the most impact with the highest SHAP values for the three crash types. Specifically, the maneuver executed by the at-fault vehicle, such as changing lanes, traveling straight, or performing other actions, directly affects the positioning and interaction between vehicles, thereby influencing the type of crash. The crash location also plays an important role with higher SHAP values of \textit{Crash\_location\_On roadway - Non-intersection} for rear-end and sideswipe crashes than angle crash. 

The traffic features, such as the hourly volume (\textit{Hourly\_volume}), hourly average speed (\textit{Hourly\_avg\_speed}), and truck percentage (\textit{Hourly\_ truck\_ratio}), see detailed explanations in Table \ref{table:dataset_description}, are listed among the top seven most important features across  crash types. These results underscore the critical role of vehicle behavior and traffic flow in affecting crash types. Similarly, hourly volume, speed, and truck ratio reflect traffic density, status, and exposure, which directly correlate with crash risks. The hourly average speed emerges as a critical feature across crash types, with notable contributions in rear-end and sideswipe crashes. Its significance can be attributed to the direct relationship between speed and the likelihood of these crashes. High average speeds reduce the reaction time available to drivers, increasing the risk of rear-end collisions, especially in high-traffic scenarios. For sideswipe crashes, elevated speeds may amplify the difficulty of safely changing lanes or maintaining control during complex driving maneuvers. The \textit{Hourly\_truck\_ratio}, which is unique in our dataset, emerges as a notable feature in the SHAP analysis, contributing to the understanding of crash type dynamics, particularly for rear-end and sideswipe collisions. This feature represents the proportion of trucks in the traffic flow during a given hour and is closely tied to crash likelihood due to the unique characteristics and limitations of trucks. The SHAP values indicate that a higher \textit{Hourly\_truck\_ratio} generally increases the probability of sideswipe crashes, likely because trucks occupy more space on the roadway, have larger blind spots, and require greater distances to maneuver safely. These factors can create conditions that make lane changes and overtaking more challenging for other vehicles, leading to sideswipe incidents. In addition, \textit{Hourly\_truck\_ratio} plays a role in rear-end crashes, albeit with a smaller magnitude of contribution. The large size, heavy weight, and slower acceleration of trucks can create speed differentials in mixed traffic, particularly during peak hours, where traffic density is high. Such conditions may increase the likelihood of rear-end collisions when vehicles following trucks fail to adjust their speed or maintain a safe following distance. These findings underscore the need for targeted interventions, such as lane restrictions for trucks during peak traffic hours, better signage to alert drivers of truck-heavy zones, and enhanced training for drivers on navigating safely around trucks. 

Additionally, other factors, including  pavement features (\textit{cross-section slope, IRI\_avg}), geometric features (\textit{Heading\_angle}), and weather-related features (\textit{gust}), provide further insights into the multifaceted nature of crash dynamics. %driver safety equipment usage (Driver1\_safety\_equip),

\subsection{FGTT attention heatmaps}

To better understand feature importance for the FGTT model, attention weights were extracted from the last attention layer and examined at different levels. In the context of attention mechanisms, there are two key elements, the query and key tokens, that play a pivotal role. These tokens are transformed representations of the input data. Specifically, query tokens are used to probe the input data, while key tokens are matched against these queries. The model evaluates the similarity between each query token and all key tokens to compute attention scores.  These scores quantify the relevance of each part of the input sequence to the query. Using these scores, the model generates a weighted sum of values, an amalgamation of contextually embedded information that highlights the features deemed most relevant for making predictions. In this study, attention weights for the FGTT model were extracted for three distinct crash types from the test set and passed through the final trained model to identify the features most associated with each crash type. The attention weights were obtained from the last layer of the transformer encoder, which is the layer that produces the final transformer output used for predictions.

\begin{figure}[H]
	\centering
	\includegraphics[width=0.8\linewidth]{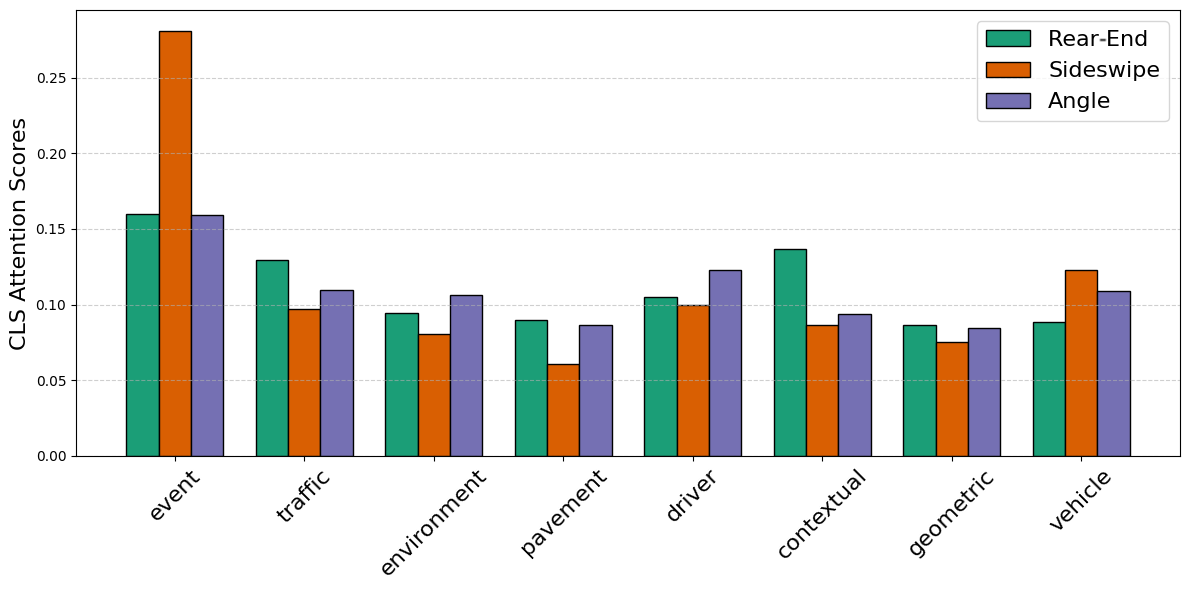}
	\caption{Class token (CLS) attention scores by feature groups.}
	\label{fig:CLS attention}
\end{figure}

Figure \ref{fig:CLS attention} illustrates the class token attention scores toward each feature group for crash type inference. These attention scores signify the importance of each feature group in predicting three crash types: rear-end, sideswipe, and angle.

The results show that the event feature group consistently receives the highest attention scores across all crash types from the class token. This indicates the dominant role of event-specific details, such as vehicle maneuvers and crash locations, affecting the risks of different crash types. Notably, sideswipe crashes exhibit higher attention to the event feature group compared to angle crashes and rear-end crashes, likely reflecting the critical influence of dynamic actions, such as lane changes, which are commonly associated with these types of collisions. In contrast, rear-end and angle crashes may depend less on immediate events and more on environmental, contextual, and pavement factors.

The traffic, environment, pavement, driver, and contextual feature groups demonstrate moderate attention scores, suggesting their relevance in providing additional context for crash type prediction. For example, angle crashes exhibit relatively higher attention to environmental features, which may highlight the role of conditions like \textit{gust}, \textit{wind speed}, \textit{precipitation}, and \textit{lighting} in angle crash scenarios, particularly on the driver's driving perception and pavement conditions. Similarly, the driver and contextual feature groups receive comparable attention across crash types, suggesting the importance of driver characteristics, potentially driver behavior indicated by the driver characteristics, and contextual factors, such as traffic patterns, reflected by the time of day and the day of the week. These features likely serve as secondary factors that modulate the primary event-driven dynamics of crashes. This finding aligns with the understanding that immediate crash events and driver interactions often have a stronger impact on outcomes than static roadway or vehicle characteristics.

\begin{figure}[htbp!]
	\centering
	\includegraphics[width=0.4\linewidth]{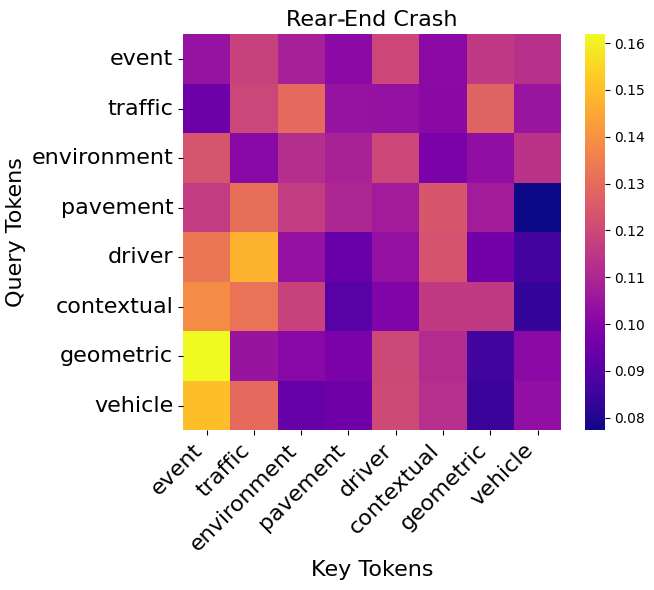}
	\caption{Feature groups attention heatmap for rear-end crashes.}
	\label{fig:Feature groups attention heat map for rear end}
\end{figure}

\begin{figure}[htbp!]
	\centering
	\includegraphics[width=0.4\linewidth]{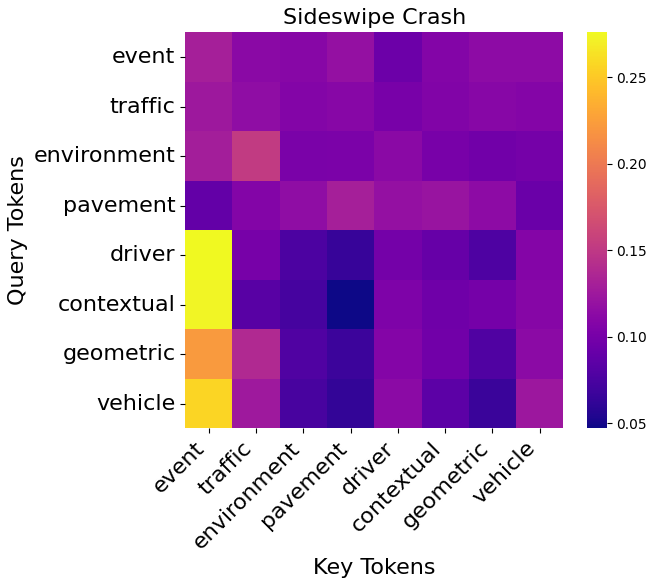}
	\caption{Feature groups attention heatmap for sideswipe crashes.}
	\label{fig:Feature groups attention heat map for sideswipe}
\end{figure}

\begin{figure}[htbp!]
	\centering
	\includegraphics[width=0.4\linewidth]{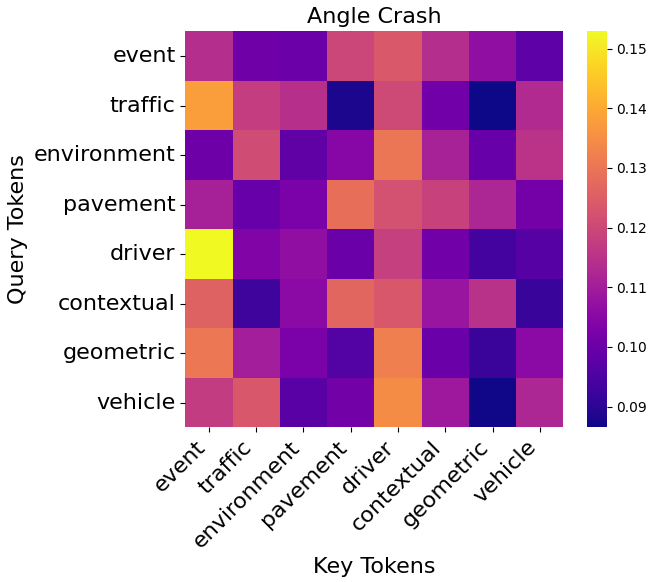}
	\caption{Feature groups attention heatmap for angle crashes.}
	\label{fig:Feature groups attention heat map for angle}
\end{figure}

Figures \ref{fig:Feature groups attention heat map for rear end}--\ref{fig:Feature groups attention heat map for angle} present the attention heatmaps depicting the interactions between different feature groups for three crash types: rear-end, sideswipe, and angle. The intensity of the attention weights in these headmaps reflects the importance of the relationship for each feature group pair in the model's predictions.

Analyzing the heatmaps reveals distinct patterns of feature group interactions for each crash type. Comparing the heatmaps across crash types reveals that certain feature group interactions are universally influential, while others are crash-type specific. One notable finding across the three heatmaps is that the event feature group shows more influence than other feature groups across the three crash types. For the rear-end crash type, there are prominent interactions between feature groups, including \textit{event} and \textit{geometric}, \textit{traffic} and \textit{driver}, \textit{event} and \textit{vehicle}, \textit{event} and \textit{contextual}, \textit{event} and \textit{driver}  feature groups. For sideswipe crash type, the heatmap highlights substantial interactions of the \textit{event} feature group with the \textit{driver}, \textit{contextual}, \textit{vehicle} feature groups. 

Moreover, the \textit{event} feature group has consistent strong attention toward the \textit{driver} feature group across all three heatmaps. The persistent interaction  underscores the fundamental role of \textit{driver}-related features, or potential latent factors, in the occurrence of different crash types. 

In the angle crash heatmap, there is more pronounced emphasis on overall feature group interactions compared to rear-end and sideswipe crashes. This  may suggest a more complex combined effect of feature group interactions on the model's inference for angle crashes. Notably, the strong attention associated with \textit{driver}-\textit{event} feature group pairs further highlights their significant influence in predicting the angle crash type.

These findings demonstrate the model's ability to capture the intricate relationships among  different feature groups in inferring crash types. The highlighted feature group interactions specific to each crash type reveal the potential for developing targeted intervention strategies. For instance, mitigating rear-end crashes could involve prioritizing driver assistance technologies and improving pavement conditions. Reducing sideswipe crashes might require redesigning roadway geometries and managing traffic flows. Addressing angle crashes could benefit from enhancing intersection designs and implementing environmental hazard warning systems.

In summary, the attention heatmaps offer valuable insights into how the transformer-based model utilizes feature group interactions to distinguish between crash types. Understanding these interactions not only enhances the interpretability of the model but also guides the development of targeted safety measures to mitigate specific crash types effectively.

\section{Conclusions and future directions}

Drawing from the collective insights gathered throughout the research, it becomes evident that predictive modeling of traffic crash outcomes benefits significantly from the integration of diverse data types and sources. The primary contributions of this study lie in the use of a comprehensive and highly descriptive dataset and the introduction of the FGTT, which efficiently processes and encodes data into group-specific tokens, enabling it to uncover complex relationships between feature groups through attention heatmaps. These semantically rich tokens enhance the model's ability to capture nuanced interactions, resulting in both improved predictive performance and greater interpretability. While the FGTT model demonstrated better performance in our experiments compared to  popular ensemble methods, opportunities for further refinement remain. In this study, the token feature groups were constructed using carefully selected features identified as semantically similar based on domain knowledge. Future research could explore  more flexible grouping schemes that  allow the model to dynamically learn optimal groupings to enhance the semantic understanding of feature interactions and further improve the model's ability to infer crash outcomes. 

Further insights were gained by examining the attention weights extracted from the FGTT model, providing a deeper understanding of the factors influencing different crash types. In particular, for angle crashes, distinct attention patterns emerged, with driver and environment features receiving greater weight compared to the other two crash types. For sideswipe and rear-end crashes, event-related features, such as crash location and vehicle maneuvers, and contextual-related features are more prominent. Moreover, the event feature group has relatively strong attention toward the driver feature group across all three crash types, underscoring the importance of driver features. Considering the growing emphasis on explainable machine learning, enhancing the interpretability of complex models like the FGTT is critical. Developing methods to make the model’s decision-making more comprehensible will not only bolster trust but also support its adoption in policy-making and real-world applications, where transparency and interpretability  are essential. 

Furthermore, the comparative evaluation of different modeling methods underscored the effectiveness of the CatBoost model in handling the minority classes, evidenced by its leading performance in predicting the angle crash category. Future research should explore SMOTE or generative models, such as generative adversarial networks or variational autoencoders, to synthetically augment minority classes with high-fidelity samples \cite{hernandez2022synthetic}. These approaches could help create a more balanced training dataset, ultimately enhancing the model’s predictive power across classes. Additionally, the exploration of alternative data augmentation techniques, innovative feature engineering strategies, and the integration of physics-guided or domain-specific embeddings could also be invaluable.

From a data perspective, in addition to CCS data utilized in this study, weigh-in-motion (WIM) data is another valuable data source than could enhance crash outcome prediction. Previous studies have identified correlations between truck weights and crash outcomes \cite{jo2019estimation,xu2023exploring}, suggesting that incorporating WIM data could improve model performance. However, a challenge in utilizing WIM data, however, is its sparse distribution, which would limit the amount of crash instances available for analysis at these WIM sites. While this study considered hourly summaries, further insights can be gained through the exploration and usage of finer time resolutions (e.g., 5 minutes or 15 minutes), as Dutta and Fontaine \cite{dutta2019improving} found that diverse ways to aggregate the traffic data had impacts on their modeling results.

The growth of the electric vehicle (EV) and connected and autonomous vehicles (CAV) present new opportunities for future research. Many new EVs on the market are equipped with an array of cameras and sensors that can provide a richer data source, offering deeper insights into driver behavior, actions, and vehicle maneuvers. In conclusion, this research represents a significant step forward in enhancing road safety analytics. By leveraging the FGTT’s ability to uncover complex relationships within the data through semantically grouped features, it lays the groundwork for more informed and effective safety strategies. Additionally, the comprehensive dataset utilized in this study underscores the importance of considering diverse features that impact crash outcomes. It is the convergence of data-driven insights and technological innovation that will drive progress toward safer roadways and more effective policy making for a safer future.

\section*{Use of AI tools declaration}
The authors declare they have not used Artificial Intelligence (AI) tools in the creation of this article.

\section*{Conflict of interest}
The authors declare no conflict of interest.

\bibliographystyle{unsrt}  
\bibliography{references}  

\end{document}